\documentclass[runningheads]{llncs}

\usepackage{eccvabbrv}
\usepackage{xcolor}
\usepackage{graphicx}
\usepackage{booktabs}

\usepackage[accsupp]{axessibility}  

\usepackage{orcidlink}

\begin{document}

\title{AniPortrait: Audio-Driven Synthesis of Photorealistic Portrait Animation} 

\titlerunning{AniPortrait}

\author{Huawei Wei*\and Zejun Yang*\and Zhisheng Wang}

\authorrunning{Wei et al.}

\institute{Tencent \\
\email{\{huaweiwei,zejunyang,plorywang\}@tencent.com}}

\footnotetext[1]{*Equal contribution.}

\maketitle

\begin{abstract}
  In this study, we propose AniPortrait, a novel framework for generating high-quality animation driven by 
audio and a reference portrait image. Our methodology is divided into two stages. Initially, we extract 3D intermediate representations from audio and project them into a sequence of 2D facial landmarks. Subsequently, we employ a robust diffusion model, coupled with a motion module, to convert the landmark sequence into photorealistic and temporally consistent portrait animation. Experimental results demonstrate the superiority of AniPortrait in terms of facial naturalness, pose diversity, and visual quality, thereby offering an enhanced perceptual experience. Moreover, our methodology exhibits considerable potential in terms of flexibility and controllability, which can be effectively applied in areas such as facial motion editing or face reenactment. We release code and model weights at \href{https://github.com/scutzzj/AniPortrait}{https://github.com/Zejun-Yang/AniPortrait.}
\end{abstract}

\section{Introduction}
\label{sec:intro}

The creation of realistic and expressive portrait animations from audio and static images has  a range of applications, spanning from virtual reality and gaming to digital media. Yet, the production of high-quality animations that are visually captivating and maintain temporal consistency presents a significant challenge. This complexity arises from the need for intricate coordination of lip movements, facial expressions, and head positions to craft a visually compelling effect.

Existing methods have often fallen short in overcoming this challenge, primarily due to their reliance on limited-capacity generators for visual content creation, such as GANs\cite{guan2023stylesync,zhang2023dinet}, NeRF\cite{ye2023geneface,ye2023geneface++}, or motion-based decoders\cite{zhang2023sadtalker,ma2023dreamtalk}. These networks exhibit limited generalization capabilities and often lack stability in generating high-quality content. Recently, the emergence of diffusion models\cite{dhariwal2021diffusion,ho2020denoising,rombach2022high} has facilitated the generation of high-quality images. Some studies have built upon this by incorporating temporal modules, enabling diffusion models to excel in creating compelling videos.

Building upon the advancements of diffusion models, we introduce AniPortrait, a novel framework designed to generate high-quality animated portraits driven by audio and a reference image. AniPortrait is divided into two distinct stages. In the first stage, we employ transformer-based models to extract a sequence of 3D facial mesh and head pose from the audio input, which are subsequently projected into a sequence of 2D facial landmarks.  This stage is capable of capturing subtle expressions and lip movements from the audio, in addition to head movements that synchronize with the audio's rhythm.
In the subsequent  stage, we utilize a robust diffusion model\cite{rombach2022high}, integrated with a motion module\cite{guo2023animatediff}, to transform the facial landmark sequence into a temporally consistent and photorealistic animated portrait. Specifically, we draw upon  the network architecture from AnimateAnyone\cite{hu2023animate}, which utilizes a potent diffusion model, Stable Diffusion 1.5, to generate fluid and lifelike videos based on a body motion sequence and a reference image.
Of particular note is our redesign of the pose guider module within this network. This modification not only maintains a lightweight design but also exhibits heightened precision in generating lip movements.

Our experimental results show the superiority of AniPortrait in creating animations with impressive facial naturalness, varied poses, and excellent visual quality.
By employing 3D facial representations as intermediate features, we gain the flexibility to modify these representations as required. This adaptability greatly enhances the applicability of our framework in domains like facial motion editing and facial reenactment. 

\begin{figure*}[htbp]
\begin{center}
\includegraphics[width=1\linewidth]{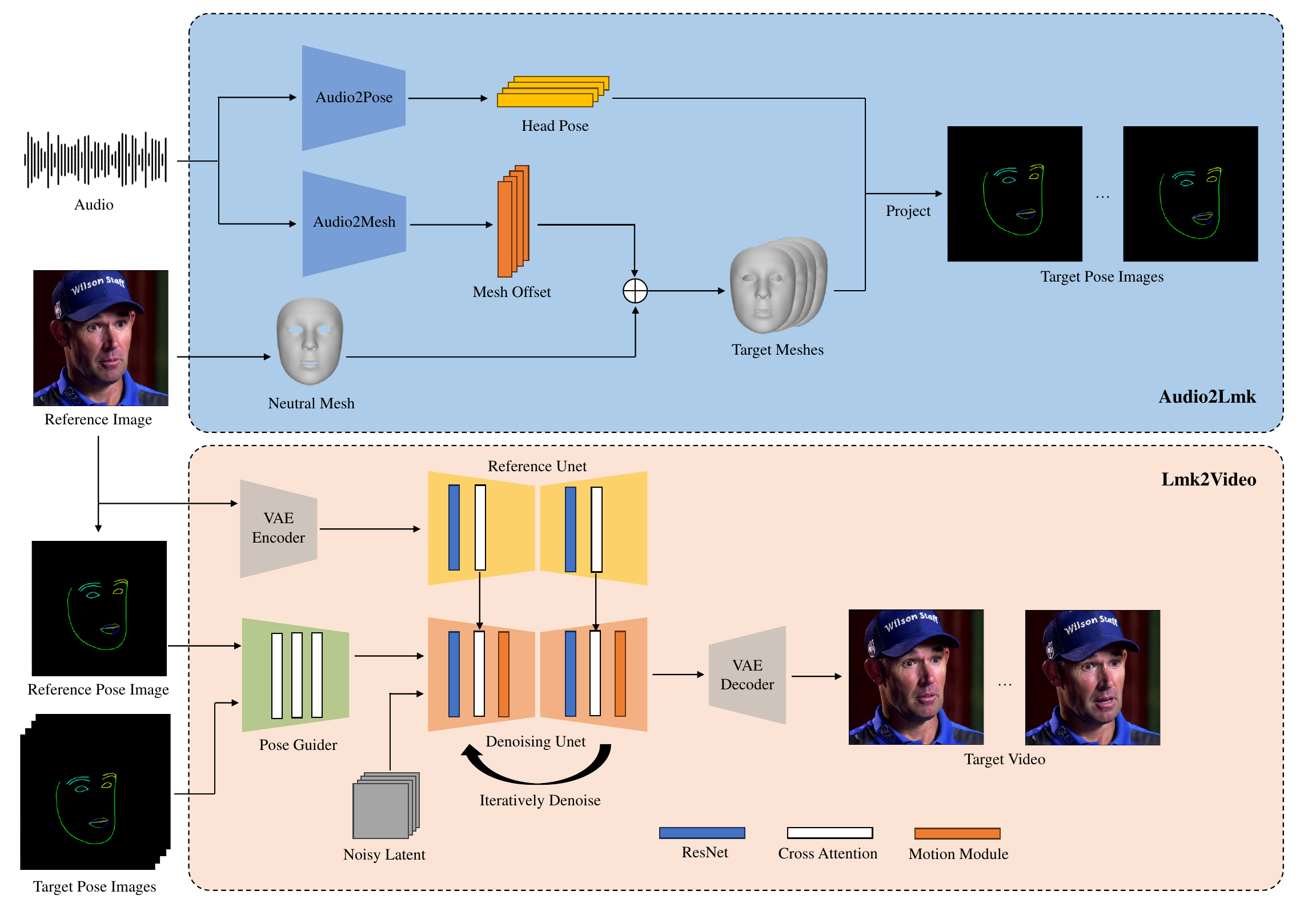}
\end{center}
\vspace{-0.8cm}
\caption{Overview of the proposed method. Our framework is divided into two stages. Firstly, we extract the 3D facial mesh and head pose from the audio, and subsequently project these two elements into 2D keypoints. In the second stage, we utilize a diffusion model to transform the 2D keypoints into a portrait video. These two stages are trained concurrently within our framework.}
\label{fig:main}
\vspace{-0.3cm}
\end{figure*}

\section{Method}
\label{sec:method}
The proposed framework comprises two modules, namely Audio2Lmk and Lmk2Video. The former  is designed to extract a sequence of landmarks that captures intricate facial expressions and lip movements from audio input. The latter  utilizes this landmark sequence to generate high-quality portrait videos with temporal stability. We present an overview of the framework in Figure \ref{fig:main} and provide further details below.

\subsection{Audio2Lmk}
\label{subsec:a2l}
Let $\mathcal{A}^{1: T}=\left(\mathbf{a}^{1}, \ldots, \mathbf{a}^{T}\right)$ denotes a sequence of speech snippets, our goal is to predict the corresponding 3D facial mesh sequence $\mathcal{M}^{1: T}=\left(\mathbf{m}^{1}, \ldots, \mathbf{m}^{T}\right)$, where each $\mathbf{m}^{t} \in \mathbb{R}^{N\times 3}$, and pose sequence $\mathcal{P}^{1: T}=\left(\mathbf{p}^{1}, \ldots, \mathbf{p}^{T}\right)$, with each $\mathbf{p}^{t}$ being a 6-dim vector representing rotation and translation.

We employ the pre-trained wav2vec\cite{baevski2020wav2vec} to extract audio features. This model exhibits a high degree of generalizability and is capable of accurately recognizing pronunciation and intonation from the audio, which plays a pivotal role in generating realistic  facial animations. By leveraging the acquired robust speech features, we can effectively employ a simple architecture consisting of two fc layers to convert these features into 3D facial meshes. We have observed that this straightforward design not only ensures accuracy but also enhances the efficiency of the inference process. 

In the task of converting audio to pose, we employ the same wav2vec network as the backbone. However, we do not share the weights with the audio-to-mesh module. This is due to the fact that pose is more closely associated with the rhythm and tone present in the audio, which is a different emphasis compared to the audio-to-mesh task.  To account for the impact of previous states, we employ a transformer\cite{vaswani2017attention} decoder to decode the pose sequence. During this process, the audio features are integrated into the decoder using cross-attention mechanisms. 
For both of the above modules, we train them using simple L1 loss. 

After obtaining the mesh and pose sequence, we employ perspective projection to transform them into a 2D sequence of facial landmarks. These landmarks are subsequently utilized as input signals for the next stage.

\subsection{Lmk2Video}
\label{subsec:l2v}
Given a reference portrait image, denoted as $ I_{ref}$, and a sequence of facial landmarks represented as  $ \mathcal{L}^{1: T}=\left(\mathbf{l}^{1}, \ldots, \mathbf{l}^{T}\right)$ where each $\mathbf{l}^{t} \in \mathbb{R}^{N\times 2}$, our proposed Lmk2Video module creates a temporally consistent portrait animation. This animation aligns the motion with the landmark sequence and maintains an appearance that is consistent with the reference image. We represent the portrait animation as a sequence of portrait frames, denoted as  $\mathcal{I}^{1: T}=\left(\mathbf{I}^{1}, \ldots, \mathbf{I}^{T}\right)$.

The design of Lmk2Video's network structure draws inspiration from AnimateAnyone. We utilize SD1.5 as the backbone, incorporating a temporal motion module that effectively converts multi-frame noise inputs into a sequence of video frames. Concurrently, a ReferenceNet, mirroring the structure of SD1.5, is employed to extract appearance information from the reference image and integrate it into the backbone. This strategic design ensures the face ID remains consistent throughout the output video. 
Difference from AnimateAnyone, we enhance the complexity of the PoseGuider's design. The original version merely incorporates a few convolution layers, after which the landmark features merge with the latents at the backbone's input layer. We discover that this rudimentary  design falls short in capturing the intricate movements of the lips. 
Consequently, we adopt ControlNet's\cite{zhang2023adding} multi-scale strategy, incorporating landmark features of corresponding scales into different blocks of the backbone. Despite these enhancements, we successfully keep the parameter count relatively low.

We also introduce an additional improvement: the inclusion of the reference image's landmark as an extra input. The PoseGuider's cross-attention module facilitates interaction between the reference landmarks and each frame's target landmarks. This process provides the network with additional cues to comprehend the correlation between facial landmarks and appearance, thereby assisting in the generation of portrait animations with more precise motion.

\section{Experiments}
\label{sec:exp}
\subsection{Implementation Details}
In the Audio2Lmk stage, we adopt wav2vec2.0 as our backbone. We leverage MediaPipe\cite{lugaresi2019mediapipe} to extract 3D meshes and 6D poses for annotations. The training data for Audio2Mesh comes from our internal dataset, which comprises nearly an hour of high-quality speech data from a single speaker. To ensure the stability of the 3D mesh extracted by MediaPipe, we instruct the actor to maintain a steady head position, facing the camera throughout the recording process. We train Audio2Pose using HDTF\cite{zhang2021flow}. All training operations are performed on a single A100, utilizing the Adam optimizer with a learning rate of 1e-5.

In the Lmk2Video process, we implement a two-step training approach. During the initial step, we focus on training the 2D component of the backbone, ReferenceNet, and PoseGuider, leaving out the motion module. In the subsequent step, we freeze all other components and concentrate on the training of the motion module. We make use of two large-scale, high-quality facial video datasets, VFHQ\cite{xie2022vfhq} and CelebV-HQ\cite{zhu2022celebv} to train the model. All data undergoes processing via MediaPipe to extract 2D facial landmarks. To enhance the network's sensitivity to lip movements, we differentiate the upper and lower lips with distinct colors when rendering the pose image from 2D landmarks. All images are resized to 512x512 resolution. We utilize 4 A100 GPUs for model training, dedicating two days to each step. The AdamW optimizer is employed, with a consistent learning rate of 1e-5.

\subsection{Results}

\begin{figure*}[htbp]
\begin{center}
\includegraphics[width=1\linewidth]{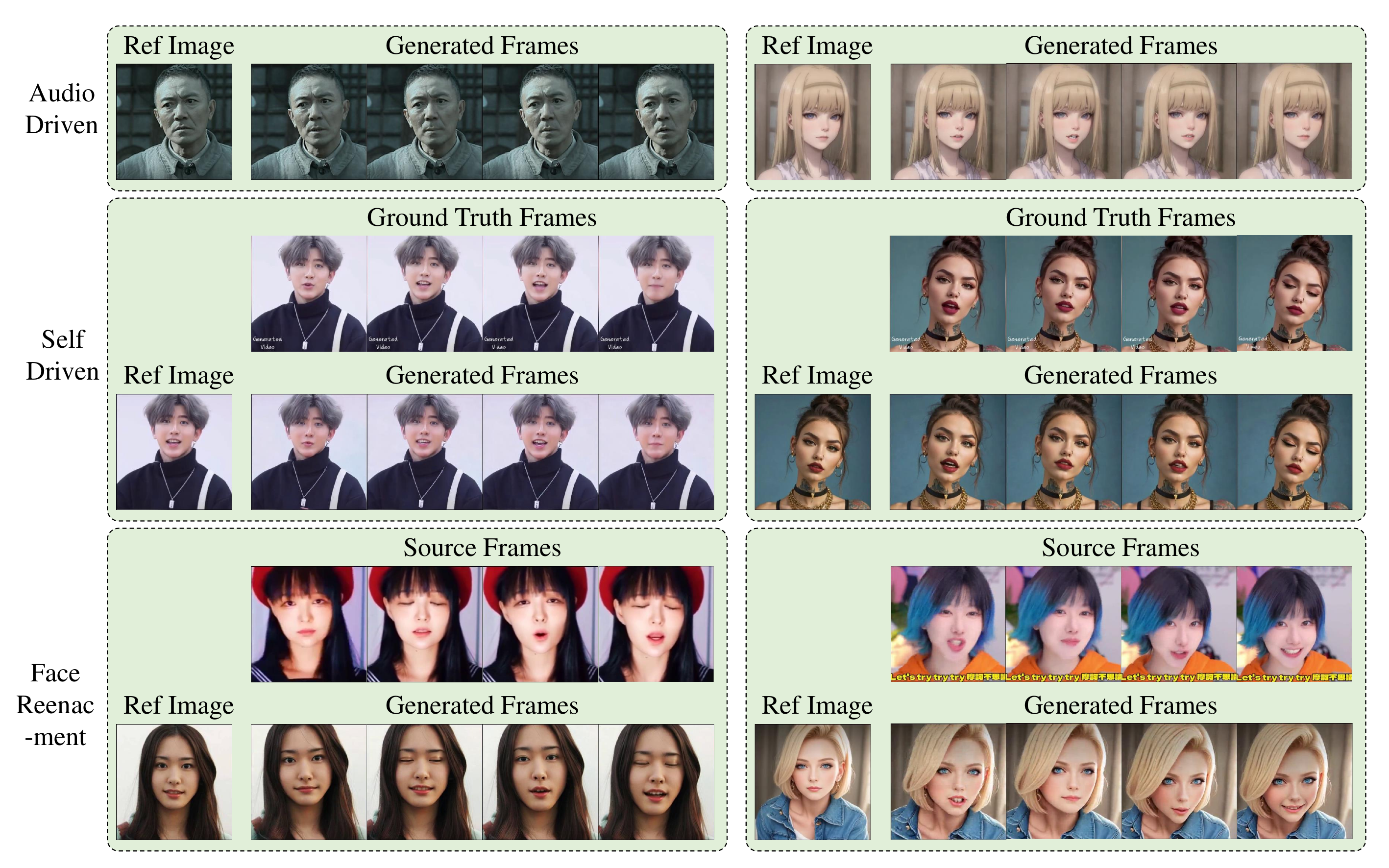}
\end{center}
\vspace{-0.8cm}
\caption{Results of our method. }
\label{fig:result}
\vspace{-0.3cm}
\end{figure*}

As depicted in Figure \ref{fig:result}, our method generates a series of animations that are striking in both quality and realism. We utilize an intermediate 3D representation, which can be edited to manipulate the final output. For instance, we can extract landmarks from a source and alter its ID, thereby enabling us to create a face reenactment effect. 

\section{Conclusion and Future Work}
This research presents a diffusion model-based framework for portrait animation. By simply inputting an audio clip and a reference image, the framework is capable of generating a portrait video that features smooth lip motion and natural head movements. Leveraging the robust generalization abilities of the diffusion model, the animations created by this framework display impressively realistic image quality and convincingly lifelike motion. However, this method necessitates the use of intermediate 3D representations, and the cost of obtaining large-scale, high-quality 3D data is quite high. As a result, the facial expressions and head postures in the generated portrait videos  can't escape the uncanny valley effect. In the future, we plan to follow the approach of EMO\cite{tian2024emo} , predicting portrait videos directly from audio, in order to achieve more stunning generation results.

%
%
\bibliographystyle{splncs04}
\bibliography{main.bib}
\end{document}